# Higher-Order DisCoCat
*(Peirce-Lambek-Montague semantics)*

Alexis Toumi and Giovanni de Felice

*Quantinuum – Quantum Compositional Intelligence*
*17 Beaumont street, OX1 2NA Oxford, UK*

November 30, 2023

## Abstract

We propose a new definition of higher-order DisCoCat (categorical compositional distributional) models where the meaning of a word is not a diagram, but a diagram-valued higher-order function. Our models can be seen as a variant of Montague semantics based on a lambda calculus where the primitives act on string diagrams rather than logical formulae. As a special case, we show how to translate from the Lambek calculus into Peirce's system beta for first-order logic. This allows us to give a purely diagrammatic treatment of higher-order and non-linear processes in natural language semantics: adverbs, prepositions, negation and quantifiers. The theoretical definition presented in this article comes with a proof-of-concept implementation in DisCoPy, the Python library for string diagrams.

## Introduction

DisCoCat [1, 2] (Categorical Compositional Distributional) models are structure-preserving maps which send *grammatical types to vector spaces* and *grammatical structures to linear maps*. Concretely, the meaning of words is given by tensors with shapes induced by their grammatical types; the meaning of sentences is given by contracting the tensor networks induced by their grammatical structure. *String diagrams* provide an intuitive graphical language to visualise and reason formally about the evaluation of DisCoCat models; which can be formalised in terms of *functors* $F : \mathbf{G} \to \mathbf{Vect}$ from the category generated by a formal grammar $\mathbf{G}$ to the monoidal category $\mathbf{Vect}$ of vector spaces and linear maps with the tensor product [3, §2.5].

Although this functorial definition applies equally to any kind of formal grammar, most of the DisCoCat literature focuses on *pregroup grammars* and more generally on *categorial grammars* such as the *Lambek calculus* [4, 5] and *combinatory categorial grammars* (CCG) [6]. In that case, $\mathbf{G}$ is a *closed monoidal category* and the DisCoCat models $F : \mathbf{G} \to \mathbf{Vect}$ map grammatical structures to the closed structure of $\mathbf{Vect}$ in a canonical way. In practice, this means that once the meaning of each word is computed from a dataset, the meaning of any new grammatical sentence can be computed automatically from its grammatical structure.

While assigning vectors to words of *atomic types* like common nouns was already a standard technique in the field of distributional semantics, the main novelty of the DisCoCat framework lied in its treatment of *complex types* like adjectives, verbs, etc. If the atomic type $n$ for common nouns is assigned a vector space $F(n) = N$, the complex type $n \leftarrow n$ for adjectives (modifying nouns on their right) is mapped to the tensor product $F(n \leftarrow n) = N \otimes N^\star$: *adjectives are not vectors, they are linear maps*. Now if we play the same game with the type for adverbs (modifying adjectives on their right) we get tensors of order four: $F((n \leftarrow n) \leftarrow (n \leftarrow n)) = N \otimes N^\star \otimes N^\star \otimes N$. Not only does this lead to unreasonably large representations — if we want 1000 parameters for each noun, we need 1000 billion parameters *for each adverb* — but we argue that it also misses the point of such higher-order types altogether.

Because the codomain of DisCoCat models $F : \mathbf{G} \to \mathbf{Vect}$ is a *compact-closed category*, any second-order type $a \leftarrow (b \leftarrow c)$ is squashed down to the same vector space $A \otimes (B \otimes C^\star)^\star \simeq A \otimes C \otimes B^\star$ as the first-order types $a(c \leftarrow b)$ and $(ac) \leftarrow b$. This collapse down to first-order makes it impossible to express linguistic phenomena that are inherently higher-order[1] such as negation, disjunction and universal quantification. Indeed, the variant of DisCoCat models valued in the compact closed category of relations **Rel** are as expressive as *regular logic*, the fragment of first-order logic generated by truth, conjunction and existential quantification [7]. The one missing ingredient to get from regular logic to all of first-order logic is *negation*, which cannot be expressed in terms of relational composition. In short, *negation cannot be a relation* but it can be expressed as a non-linear, higher-order function which maps relations to relations.

This observation can be traced back to Peirce [8]'s *system beta* and his *existential graphs* which are both the first examples of string diagrams and arguably the first axiomatisation of first-order logic. In today's terminology, Peirce defined string diagrams with *swaps*, *cups*, *caps*, *spiders* — which Peirce called *lines of identity* — and a *unary operator on homsets* for negation represented as a circle that he called *cut*. Existential graphs come with graphical equations that are sound and complete for first-order logic, see [9, 10] for a categorical presentation.

The use of higher-order functions for natural language semantics is the basis of *Montague semantics* [11]. Concretely, the meaning of a word is given by a term in the simply-typed lambda calculus with the symbols of first-order logic[2] taken as primitives; the meaning of a sentence is given by composing these lambda terms together to get a closed formula. Abstractly, this can be formalised as a closed monoidal functor $F : \mathbf{G} \to \Lambda L$ where $\mathbf{G}$ is a categorial grammar and $L$ is the language of first-order logic (predicates, negation, conjunction, quantifiers, etc.) seen as the presentation of a cartesian closed category $\Lambda L$ [3, §2.2]. Interpretations of the logic in a model then induce closed monoidal functors $[\![-]\!] : \Lambda L \to \mathbf{Set}$, which we can compose with $F$ to get a set-theoretic interpretation of the grammar $[\![-]\!] \circ F : \mathbf{G} \to \mathbf{Set}$.

Let us summarise the situation. On the one hand, we have DisCoCat models defined in terms of tensors that can be learnt from data and compared efficiently (with e.g. inner products) but that cannot express higher-order phenomena such as negation and universal quan-tification. On the other, we have the higher-order functions of Montague semantics, but where the resulting logical formulae are not easily amenable to machine learning and computationally intractable (e.g. equivalence of first-order logic formulae is undecidable). What we propose is to get the best of both worlds. As in Montague semantics, we assign a simply-typed lambda term to each word, but now with diagram-valued operations as primitives. The meaning of a sentence is given by composing these lambda terms together to get a diagram which can be interpreted in **Vect**, as in a DisCoCat model, or in fact in any monoidal category **C**.

---

[1] Here we mean higher-order in the sense of functions taking functions as input, although we are still within the realm of first-order logic with quantifiers over variables rather than predicates. Note that there is also a third ambiguous meaning of order: an element of $A \otimes C \otimes B^\star$ is called a tensor of order three but it is still a first-order function when seen as a linear map $A \otimes C \to B$.

[2] Montague semantics dealt with first-order *modal logic*, we leave the treatment of modalities in natural language using Peirce's system gamma as a promising direction for future work.

We propose to go from standard DisCoCat models $F : \mathbf{G} \to \mathbf{Vect}$ to *higher-order DisCoCat* (HO-DisCoCat) models defined as closed monoidal functors $F : \mathbf{G} \to \Lambda D$ where $D$ is the language of *string diagrams* (boxes, identity, composition, tensor, etc.) again seen as the signature for a free cartesian closed category $\Lambda D$. The closed monoidal functors $[\![-]\!] : \Lambda D \to \mathbf{Set}$ now correspond precisely to interpretations of the diagrams in any monoidal category **C**. Thus we can replace **Vect** by any monoidal category **C**, which need not be compact closed. This allows many new variants of DisCoCat models, for example based on neural networks or stochastic processes. It also has potential applications in quantum natural language processing [12] (QNLP) where the compact closed structure comes at the cost of exponentially low probability of success [13].

The rest of this article is organised as follows. Section 1 gives a first toy example of a higher-order process which cannot be accounted for in these standard DisCoCat models. Section 2 formulates a standard example of Montague semantics in terms of a closed monoidal functor $F : \mathbf{G} \to \Lambda L$. Section 3 contains our main definition: HO-DisCoCat models as closed monoidal functors $F : \mathbf{G} \to \Lambda D$. Section 4 shows how our models can be extended to monoidal categories with additional operations, which we use to model arbitrary adverbs depicted as *boxes with holes* that can hold diagrams themselves. Section 5 defines a HO-DisCoCat model which we call *Peirce-Lambek-Montague* semantics, where the diagrams are existential graphs. This closes the conjecture opened in [1]: DisCoCat models valued in **Rel** are not as expressive as Montague semantics, but higher-order DisCoCat models are. Appendix A showcases an implementation of this Peirce-Lambek-Montague semantics using DisCoPy [14], the Python toolkit for computing with string diagrams.

## 1 A toy example: "very big" = "big big"

To keep things concrete, we delay the formal definition of HO-DisCoCat models and start with a simple example of a higher-order function: the function $\mathtt{twice} : (X \to X) \to (X \to X)$ which takes a function and applies it twice, i.e. $\mathtt{twice} = \lambda f x . f f x$. In our toy model of semantics:

- the noun "car" is sent to some vector $F(\mathrm{car}) \in N$ drawn as a box with one wire out,

- the adjective "big" is sent to the function which takes a vector and post-compose it with some linear map $F(\mathrm{big}) = \lambda x . B \circ x$ drawn as a box with one wire in and one wire out,

- the adverb "very" is sent to the non-linear, higher-order function $F(\mathrm{very}) = \mathtt{twice}$.

We can now apply it to the noun phrase "very big car" and check that indeed it has the same semantics as "big big car". The image of the functor $F : \mathbf{G} \to \Lambda D$ on a syntax tree can be visualised by annotating each node with a lambda term with string diagrams as primitives. The leaves of the tree contain the lambda terms for each word. Each binary node is the application of a subtree of type $x \leftarrow y$ on the left ($y \to x$ on the right) to one of type $y$ on the other side.

**Remark.** *The following notation for syntax trees annotated with their semantics is borrowed from the one used for CCG derivations [15], except that we use string diagrams rather than the traditional predicate-argument structure. Note that these are not quite string diagrams in the category* **Set***: boxes are not labeled with their corresponding function, but with the result of applying this function to its arguments.*

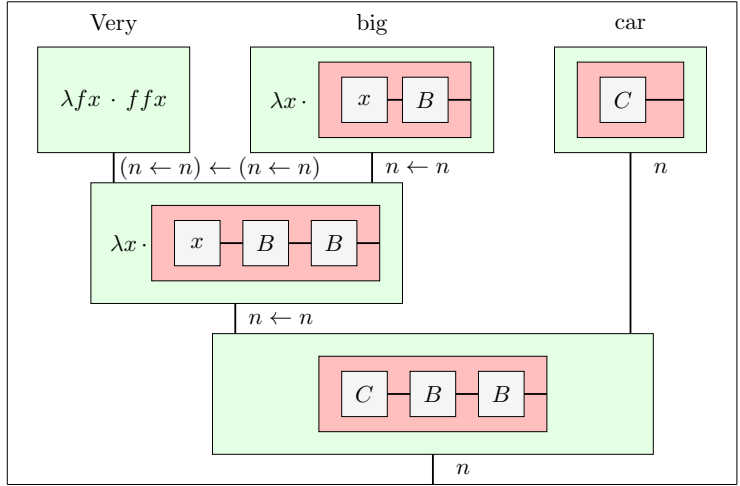

Due to the no-cloning theorem, copying an adjective in this way is not possible in a standard DisCoCat model $F : \mathbf{G} \to \mathbf{Vect}$. Indeed, there can be no linear map $\texttt{twice} : X \otimes X^\star \to X \otimes X^\star$ such that $\texttt{twice}(f) = f \circ f$ for all linear maps $f : X \to X$. How can our higher-order DisCoCat model $F : \mathbf{G} \to \Lambda D$ get away with such a fundamental law of physics? To clear any such confusion, it is important to distinguish the three categories at play in the example above:

1. The (white) closed monoidal category $\mathbf{G}$ generated by arrows $w \to t$ for each word $w$ of type $t$. The arrows $w_1 \ldots w_n \to s$ are the syntax trees of grammatical sentences.

2. The (green) cartesian closed category $\Lambda D$ where the arrows are lambda terms with the composition of string diagrams as primitives. Thanks to the cartesian product, lambda terms can copy and discard variables at will.

3. The (pink) monoidal category **Vect** where the computation of the model happens. This can be replaced by any monoidal category $\mathbf{C}$, be it cartesian, compact or otherwise.

**Remark.** *The equivalence between the lambda calculus with simple types and cartesian closed categories is known as the Curry-Howard-Lambek isomorphism [16, §1.6]. The correspondence between categorial grammars and closed monoidal categories is due to Lambek [17].*

Thus, higher-order DisCoCat models allow to make a distinction between the resource-sensitive world of **Vect** where computation happens and the cartesian category $\Lambda D$ where we can do arbitrary meta-computation. This may summarised in a slogan: *processes cannot copy, but we can copy processes*. Take the example of quantum circuits: there is no circuit that can copy arbitrary qubits, but we can copy the classical instructions for any quantum circuit.

## 2  Montague semantics as a closed monoidal functor

Before we define HO-DisCoCat and its string-diagram-valued lambda calculus, we first look at the simpler case of standard Montague semantics, defined as a monoidal functor $F : \mathbf{G} \to \Lambda L$. We adapt the presentation from [18].

As codomain $\Lambda L$, we take the cartesian closed category with objects generated by $L_0 = \{\tau, \phi\}$ for *terms* and *formulae*. The generating arrows $L_1$ are listed in the table below, they are chosen so that the arrows $f : \tau^{\otimes k} \to \phi$ in $\Lambda L$ are all the well-formed formulae of first-order logic with $k \geq 0$ free variables. The relations $(f, f') \in L_2$ between pairs of formulae $f, f' : \tau^{\otimes k} \to \phi$ may be given by any sound and complete deduction system for first-order logic, so that $f = f'$ in $\Lambda L$ if and only if $f$ and $f'$ are equivalent formulae. The closed monoidal functors $[\![-]\!] : \Lambda L \to \mathbf{Set}$ are *interpretations*, also called first-order *structures*: they are defined by a set $[\![\tau]\!]$ called the *universe* and a set $[\![\phi]\!]$ of *truth values*, together with an element $[\![\text{Alice}]\!] \in [\![\tau]\!]$ for each constant, a function $[\![\text{mortal}]\!] : [\![\tau]\!] \to [\![\phi]\!]$ for each unary predicate, etc.

| $L_1$ | symbol | type |
|---:|:---:|:---|
| **constants** | Alice, Bob, ... | $\tau$ |
| **unary predicates** | man, sings, ... | $\tau \to \phi$ |
| **binary predicates** | needs, sees, ... | $\tau \to (\tau \to \phi)$ |
| **nullary operators** | $\top, \bot$ | $\phi$ |
| **unary operators** | $\neg$ | $\phi \to \phi$ |
| **binary operators** | $\wedge, \vee, \to$ | $\phi \to (\phi \to \phi)$ |
| **quantifiers** | $\forall, \exists$ | $(\tau \to \phi) \to \phi$ |

**Remark.** *We treat the quantifiers $\forall, \exists$ as higher-order functions $(\tau \to \phi) \to \phi$ and write $\exists x.u$ for $\exists(\lambda x.u)$. In other words, we define first-order logic formulae as lambda terms where every $\lambda$ is preceded by $\forall$ or $\exists$. Thus, the variables of first-order logic formulae are encoded as variables in lambda terms, with implicit $\alpha$-reductions so that e.g. $(\exists x.Px) = (\exists y.Py)$.*

**Remark.** *The presentation $L$ implicitly depends on a* first-order signature *which contains all the non-logical symbols (constants and predicates).*

As domain **G**, we take the free closed monoidal category with objects generated by the grammatical types $\{s, n, p\}$ for *sentence*, *common noun* and *noun phrase* respectively, together with the words in some finite *vocabulary* $w \in V$. There is a generating arrow $w \to t$ for each *dictionary entry*, i.e. for each word $w \in V$ of type $t$ as listed in the table below. Now we can define the image of our functor on generating objects: sentences go to formulae, common nouns go to predicates and noun phrases go to functions from predicates to formulae. For example, the function for the noun phrase "no man" takes as input any formula $P(x)$ with a free variable and return the closed formula $\forall x.\neg \text{man}(x) \vee P(x)$.

$$F(s) = \phi \qquad F(n) = \tau \to \phi \qquad F(p) = (\tau \to \phi) \to \phi$$

We assume that the objects for each word go to the monoidal unit, i.e. $F(w) = 1$ so that the meaning of dictionary entries $w \to t$ is given states $F(w \to t) : 1 \to F(t)$ listed below.

| Montague semantics | **word** $w$ | **type** $t$ | **meaning** $F(w \to t)$ |
|---:|:---:|:---:|:---:|
| **proper nouns** | Alice | $p$ | $\lambda P.P(Alice)$ |
| **common nouns** | man | $n$ | $\lambda x.\text{man}(x)$ |
| **adjectives** | big | $n \leftarrow n$ | $\lambda Px.\text{big}(x) \wedge Px$ |
| **determiners** | every | $p \leftarrow n$ | $\lambda PQ.\forall x.Px \to Qx$ |
| **intransitive verbs** | sleeps | $p \to s$ | $\lambda P.P(\lambda x.\text{sleeps}(x))$ |
| **transitive verbs** | kills | $(p \to s) \leftarrow p$ | $\lambda PQ.Q(\lambda x.P(\lambda y.\text{kills}(x,y)))$ |

Again, we can visualise the action of the functor $F : \mathbf{G} \to \Lambda L$ on the syntax trees for a grammatical sentence by labeling each node with a lambda term. Note how the meanings of determiners "every" and "a" use both the higher-order and cartesian structure of the cartesian closed category $\Lambda L$: they take two predicates $P$ and $Q$, quantify over a variable $x$ of which they make two copies, feeding one to $P$ and the other to $Q$ before combining the resulting formulae.

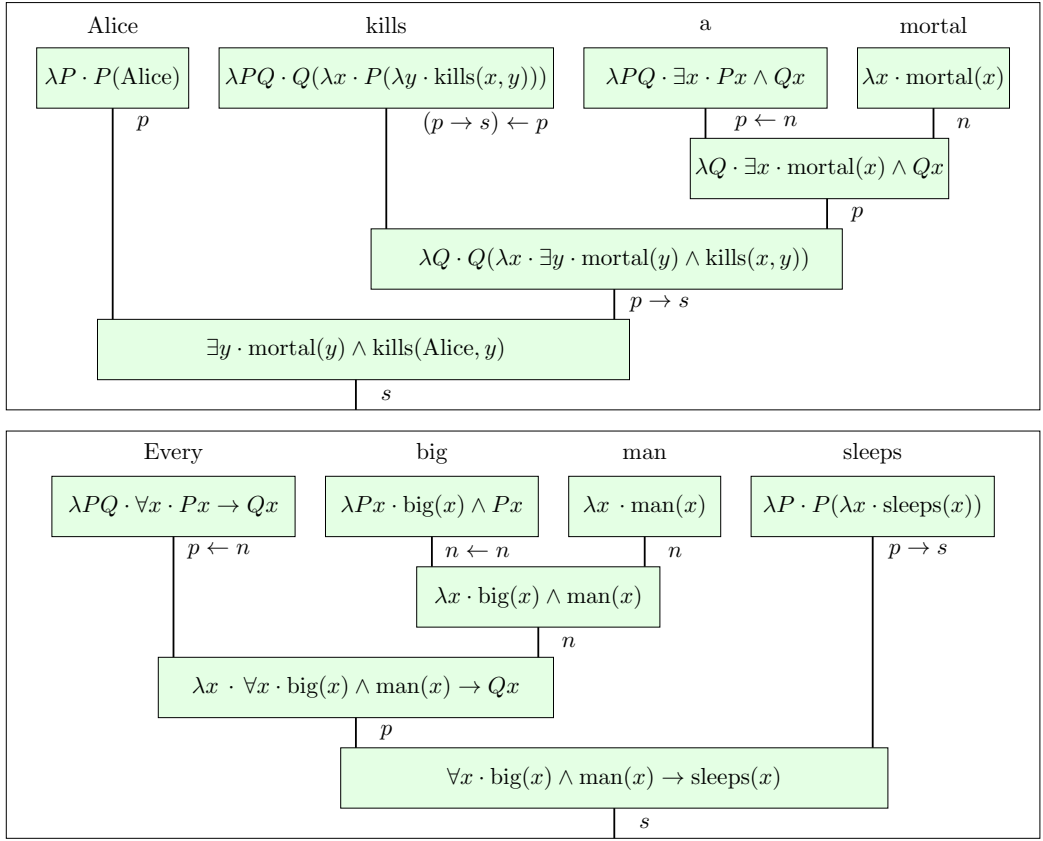

# 3 Higher-order DisCoCat as a closed monoidal functor

We can now unpack our main definition: HO-DisCoCat models as closed monoidal functors $F : \mathbf{G} \to \Lambda D$ from a categorial grammar $\mathbf{G}$ to the cartesian closed category $\Lambda D$ generated by $D$, the language of *string diagrams*.

Before we draw any diagram, we fix a monoidal signature $\Sigma = (\Sigma_0, \Sigma_1, \texttt{dom}, \texttt{cod})$, i.e. a pair of sets $\Sigma_0$ and $\Sigma_1$ for generating *objects* and *boxes* together with a pair of functions $\texttt{dom}, \texttt{cod} : \Sigma_1 \to T$ from boxes to lists of objects $T = \Sigma_0^\star = \coprod_{k \in \mathbb{N}} \Sigma_0^k$, also called *types*. The generating objects of the cartesian closed category $\Lambda D$ are given by pairs of types which encode the input and output of diagrams, i.e. $D_0 = T \times T$.

| $D_1$ | symbol | type |
|---|---|---|
| **boxes** | $f \in \Sigma$ | $(x, y)$ |
| **identity** | $\text{id}_x$ | $(x, x)$ |
| **composition** | $\circ_{xyz}$ | $(x, y) \to (y, z) \to (x, z)$ |
| **tensor** | $\otimes_{xyzw}$ | $(x, y) \to (z, w) \to (xz, yw)$ |

The generating arrows $D_1$ are listed in the table above, they are chosen so that the arrows:

$$f : (z_1, w_1) \times \cdots \times (z_k, w_k) \to (x, y) \quad \in \Lambda D$$

are all the well-typed string diagrams with $(x, y) \in D_0$ as domain, codomain and $k$ free variables for boxes of shape $(z_i, w_i) \in D_0$. The relations $(f, f') \in D_2$ between pairs of diagrams are given by the axioms for monoidal categories so that $f = f'$ in $\Lambda D$ if and only if the two diagrams are equivalent (i.e. planar isotopic), see [19, Theorem 3.1]. Closed monoidal functors $[\![-]\!] : \Lambda D \to \mathbf{Set}$ are in one-to-one correspondence with monoidal functors $[\![-]\!] : F(\Sigma) \to \mathbf{C}$ from the free monoidal category generated by $\Sigma$ into any monoidal category $\mathbf{C}$. In other words, they are defined by the following data:

- an inclusion $\Sigma_0 \hookrightarrow \mathbf{C}_0$ and a morphism $[\![f]\!] \in \mathbf{C}(x, y)$ for each box $(f : x \to y) \in \Sigma$,
- a *homset* $\mathbf{C}(x, y) = [\![(x, y)]\!]$ between every pair of types,
- an *identity morphism* $[\![\text{id}_x]\!] \in \mathbf{C}(x, x)$ for each type $x \in T$,
- a *composition function* $[\![\circ_{xyz}]\!] : \mathbf{C}(x, y) \to \mathbf{C}(y, z) \to \mathbf{C}(y, z)$,
- a *tensor function* $[\![\otimes_{xyzw}]\!] : \mathbf{C}(x, y) \to \mathbf{C}(z, w) \to \mathbf{C}(xz, yw)$.

We can recover the standard definition of DisCoCat models in terms of pregroups by taking the HO-DisCoCat model which sends the syntax trees in a categorial grammar to their corresponding pregroup diagrams, which come with an implicit interpretation in $\mathbf{C} = \mathbf{Vect}$. This is based on a functor from closed to *rigid monoidal categories* which appears in the CCG version of the DisCoCat framework [6].

| CCG-to-DisCoCat | **word** $w$ | **type** $t$ | **meaning** $F(w \to t)$ |
|---|---|---|---|
| **proper nouns** | Alice, Bob | $n$ | Alice, Bob $\in \Sigma$ |
| **transitive verbs** | loves | $(n \to s) \leftarrow n$ | $\lambda fg.(\text{cup}_N \otimes \text{id}_S \otimes \text{cup}_N) \circ (g \otimes \text{loves} \otimes f)$ |

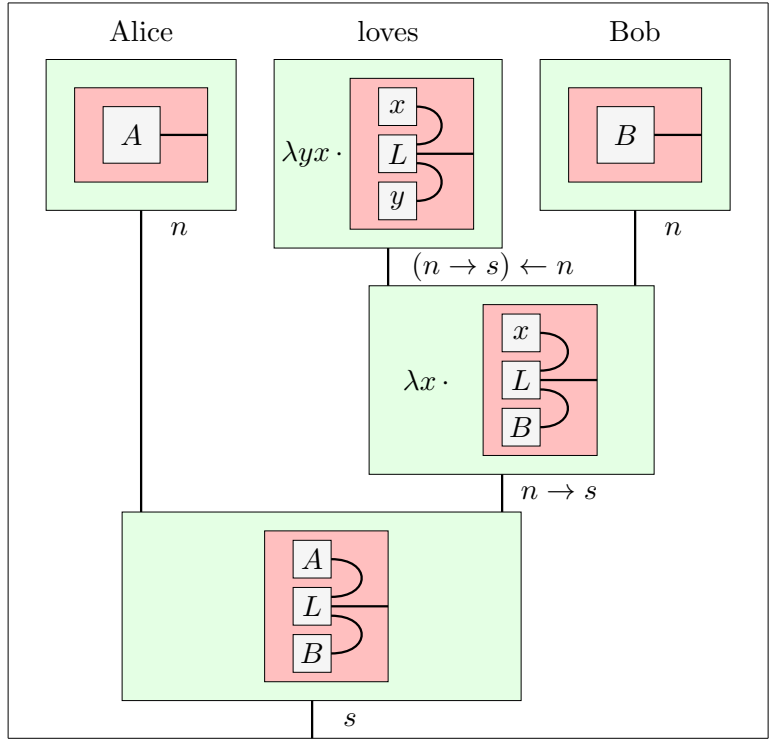

**Remark.** *A similar trick of interpreting syntax trees as functions that return pregroup diagrams is also used in the formalisation of Wittgenstein's language games in terms of functors [20].*

We can also formalise our toy example "very big car" where we define the image of our HO-DisCoCat model $F : \mathbf{G} \to \Lambda D$ as follows.

| "very big" = "big big" | **word** $w$ | **type** $t$ | **meaning** $F(w \to t)$ |
|---:|:---:|:---:|:---:|
| **common nouns** | car | $n$ | car $\in \Sigma$ |
| **adjectives** | big | $n \leftarrow n$ | $\lambda x.\text{big} \circ x$ |
| **adverbs** | very | $(n \leftarrow n) \leftarrow (n \leftarrow n)$ | $\lambda fx.ffx$ |

Note how we have three different directions in which to compose diagrams:

1. *left-to-right* using the composition $\circ_{xyz} : (x,y) \times (y,z) \to (x,z)$ which connects the output of one diagram to the input of another,

2. *top-to-bottom* using the tensor $\otimes_{xyzw} : (x,y) \times (z,w) \to (xz, yw)$ which concatenates two diagrams side by side,

3. *inside-out* using the evaluation $(x,y) \times \big[(x,y) \to (z,w)\big] \to (z,w)$ which substitutes one diagram for the free variable in another.

Whereas standard DisCoCat model could play only with the first two directions, it is this extra direction of inside-out composition that makes our higher-order DisCoCat models more flexible. In the next section, we make this extra flexibility more explicit by adding it directly to the signature for our diagrams.

# 4 Adverbs as operators on homsets

Moving away from our toy example to a more realistic model of language, we cannot assume that adverbs can all be given an explicit formula like the one we chose for "very". We propose to represent such second-order[3] processes as *boxes with k holes* that can contain diagrams of a certain shape. Concretely, we define a *monoidal signature with holes*:

$$\Sigma = (\Sigma_0, \Sigma_1, \texttt{dom}, \texttt{cod}, \text{holes})$$

as a monoidal signature together with a function $\text{holes} : \Sigma_1 \to (T \times T)^\star$ which assigns to each box $(f : x \to y) \in \Sigma$ a list of pairs of types:

$$\text{holes}(f) = \big((z_1, w_1), \ldots, (z_k, w_k)\big) \in (T \times T)^\star$$

for the domain and codomain of the diagrams that can fit in its $k \in \mathbb{N}$ holes. This is drawn as follows:

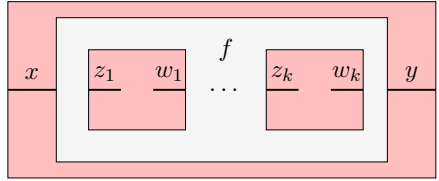

We may now add some dedicated symbols to our set of generators for the cartesian closed category $\Lambda D$, i.e. the primitives $D_1$ of our lambda calculus for diagrams.

| $D_1$ (continued) | symbol | type |
|---|---|---|
| | ... | |
| **boxes with holes** | $f \in \Sigma$ | $(z_1, w_1) \to \cdots \to (z_k, w_k) \to (x, y)$ |

The arrows $f : (z_1, w_1) \times \cdots \times (z_k, w_k) \to (x, y)$ in $\Lambda D$ are all the well-formed string diagrams where the boxes themselves can contain well-formed string diagrams, again with $k$ boxes as free variables. The closed monoidal functors $[\![-]\!] : \Lambda D \to \mathbf{Set}$ are precisely the monoidal categories $\mathbf{C}$ with a $k$-ary *operator on homsets* for each box $f \in \Sigma$:

$$[\![f]\!] : \mathbf{C}(z_1, w_1) \times \cdots \times \mathbf{C}(z_k, w_k) \to \mathbf{C}(x, y)$$

**Remark.** *A box with no holes is just a simple box; a nullary operator on homsets is just an element of a homset, i.e. a morphism. Boxes with one hole and unary operators on homsets have also been called* bubbles, *they were first introduced in [21] to denote the derivative of diagrams. As we will see in the next section, they also allow to formalise negation in Peirce's system beta. Boxes with a list of holes first appeared under the name of* quantum combs *in [22]. These "open diagrams" have then been formalised in terms of the* coend calculus *[23, 24].*

---

[3]We could imagine going higher still and consider boxes with holes that can fit diagrams with holes themselves, i.e. operators on operators on homsets. We stop at second-order because it is the limit of what we can draw with two-dimensional diagrams.

We can now define the semantics of adverbs like "furiously" as boxes with one hole for a verb, prepositions like "with" as boxes with two holes for nouns, etc.

| HO-DisCoCat | **word** $w$ | **type** $t$ | **meaning** $F(w \to t)$ |
|---:|:---:|:---:|:---|
| **nouns** | ideas | $n$ | $I \in \Sigma$ |
| **intransitive verbs** | sleep | $n \to s$ | $\lambda x. S \circ x$ |
| **adverbs** | furiously | $(n \to s) \to (n \to s)$ | $\lambda f x. F(fx)$ |
| **prepositions** | with | $(n \to n) \leftarrow n$ | $\lambda xy. W(y, x)$ |

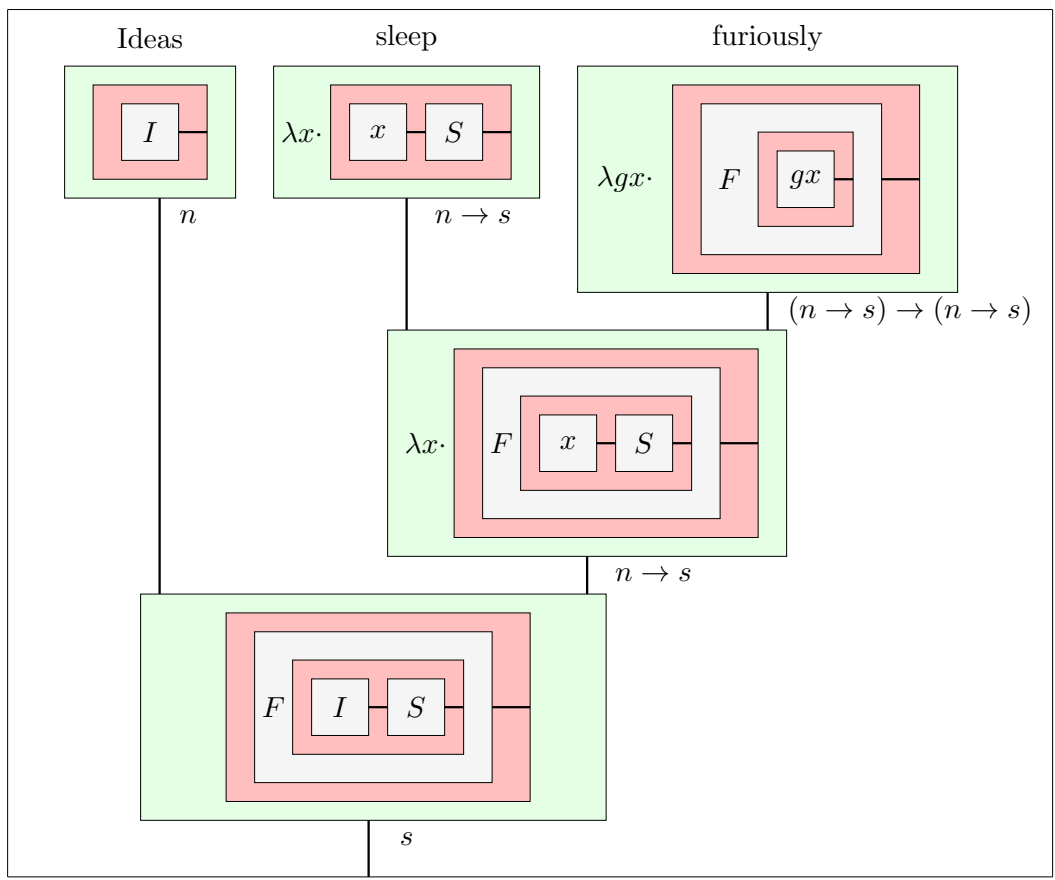

**Remark.** *Note how in the following example "concepts with attitude", we are merely interpreting the syntax tree as itself, represented in terms of nesting rather than branching. Indeed when a diagram has no composition or tensor, but only inside-out nesting of boxes with holes, it is nothing more than a syntax tree with operators on the nodes and homsets on the edges.*

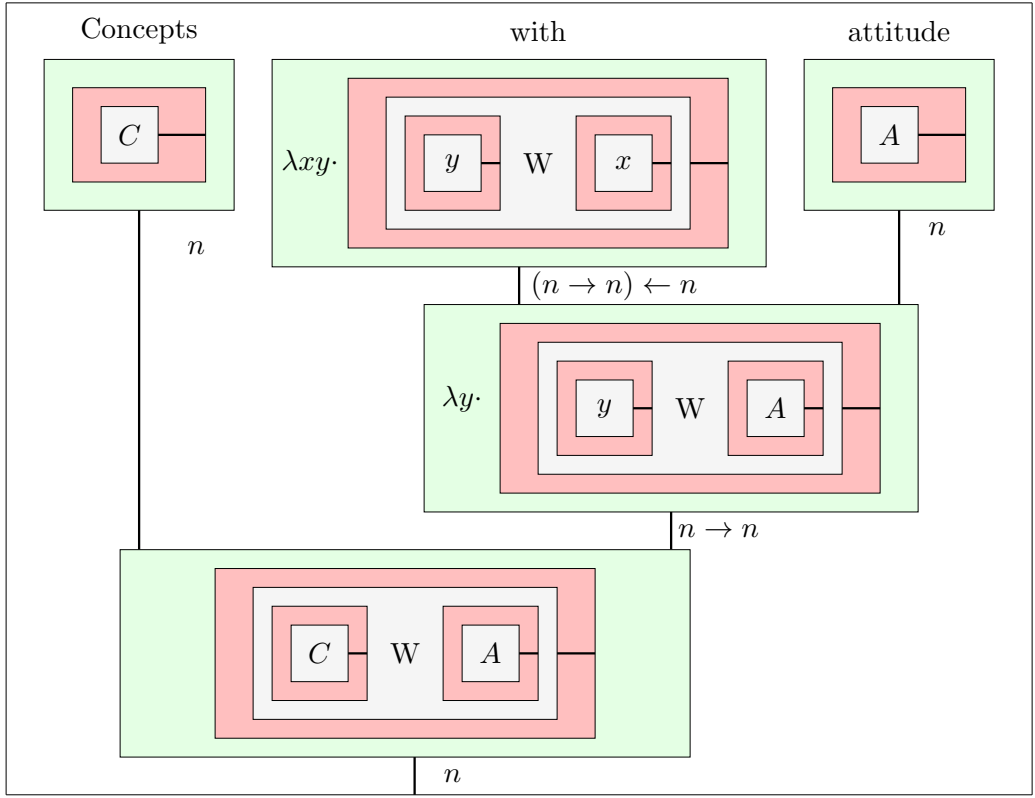

**Remark.** *Composition and tensor can themselves be thought of as binary operators on homsets, subject to the axioms of monoidal categories:* boxes with holes are all you need! *This observation can be formalised in terms of* operads of wiring diagrams *[25]*.

## 5 Peirce-Lambek-Montague semantics

In generalising Montague semantics and DisCoCat models, we are in fact going back to their common ancestor: Peirce's *existential graphs*, a formal graphical language of string diagrams for logic. Peirce's existential graphs come in three variants of increasing expressive power:

- *System alpha*, the fragment of system beta with no wires and only scalar boxes, is equivalent to propositional logic and Boolean algebras.

- *System beta* is the variant we use in this work, it is equivalent to first-order logic.

- *System gamma* extends system beta with a second unary operator on homsets for the "contingent" modality (i.e. "not necessary") together with axioms that make it sound and complete for various kinds of modal logics.

The translation between Peirce's graphical notation and the traditional one is straightforward. Boxes are predicates and wires are variables, which are implicitly existentially quantified.

Tensor is interpreted as conjunction and spiders as equality of variables. Thus, a scalar diagram with no input or output wires represents a first-order logic formula with no free variables, which we take as our interpretation for sentences. A diagram with not input and one output wire $N$ is a formula with one free variable. We take this as our interpretation for common nouns, i.e.

$$F(s) = (1,1) \qquad \text{and} \qquad F(n) = (1, N)$$

Before we move on to noun phrases and determiners, let us look at a preliminary example taken from Big Shaq's novelty song title "Man's Not Hot", which we interpret as the formula $\exists x.\mathrm{man}(x) \wedge \neg \mathrm{hot}(x)$.

| Man's Not Hot | **word** $w$ | **type** $t$ | **meaning** $F(w \to t)$ |
|---|---|---|---|
| **common noun** | Man's | $n$ | $(\mathrm{man} : 1 \to N) \in \Sigma$ |
| **negation** | Not | $(n \to s) \leftarrow a$ | $\lambda fg.(\neg f) \circ g$ |
| **adjective** | Hot | $a$ | $(\mathrm{hot} : N \to 1) \in \Sigma$ |

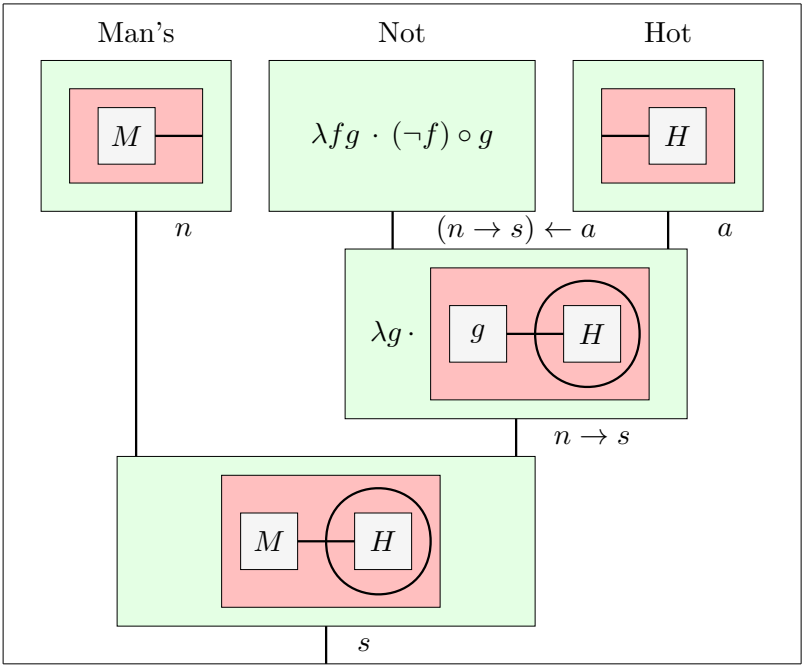

Here we had to define the type $a$ for adjectives as atomic so that we could take $F(a) = (N, 1)$ instead of $F(n \to s) = (1, N) \to (1, 1)$. Indeed, the unary operator for negation can act only on diagrams, not diagram-valued functions. Suppose we had defined the meaning of "hot" in terms of post-composition $\lambda f.\mathrm{hot} \circ f$ as in our previous example. Then the function for "not" would have had to extract the diagram $(N, 1)$ from the diagram-valued function $(1, N) \to (1, 1)$. The only way to do this would be to apply the function $\lambda x.\mathrm{hot} \circ x$ to the identity diagram $\mathrm{id}_N : (N, N)$, but that would be forbidden by its type. We need *dependent types* to express the fact that post-composition by the effect $\mathrm{hot} : (N, 1)$ can take as input any diagram $(x, N)$ and return a diagram of type $(x, 1)$.

This is precisely the definition we use to model noun phrases, i.e.

$$F(p) = \prod_x (N, x) \to (1, x)$$

For example, the function $\lambda f.\text{man} \circ f$ for the noun phrase "a man" can take as input any diagram $(N, x)$ and return a state $(1, x)$. In this way, we can apply it to an effect hot : $(N, 1)$ and get a closed formula $\exists x.\text{man}(x) \wedge \text{hot}(x)$, but we can also keep the variable free by applying it to the identity. Finally, we can define the meaning $\lambda fg.g(f(\text{id}_N)^T)$ of the copula "is": it takes two noun phrases, applies one to the identity diagram and transposes the result before feeding it to the other. Other verbs are given in the same way, replacing the identity diagram $\text{id}_N$ by any predicate.

| Peirce-Lambek-Montague | word $w$ | type $t$ | meaning $F(w \to t)$ |
|---:|:---:|:---:|:---|
| **common nouns** | man | $n$ | $(\text{man} : 1 \to N) \in \Sigma$ |
| **proper nouns** | Alice | $p$ | $\lambda f.\text{Alice} \circ f$ |
| **adjectives** | big | $n \leftarrow n$ | $\lambda f.\text{spider}_{2,1} \circ (\text{big} \otimes f)$ |
| **determiners** | no | $p \leftarrow n$ | $\lambda fg.\text{cut}(g \circ f)$ |
| **intransitive verbs** | sleeps | $p \to s$ | $\lambda P.P(\text{sleeps})$ |
| **transitive verbs** | kills | $(p \to s) \leftarrow p$ | $\lambda PQ.Q(P(\text{kills})^T)$ |
| **copula** | is | $(p \to s) \leftarrow p$ | $\lambda PQ.Q(P(\text{id}_N)^T)$ |

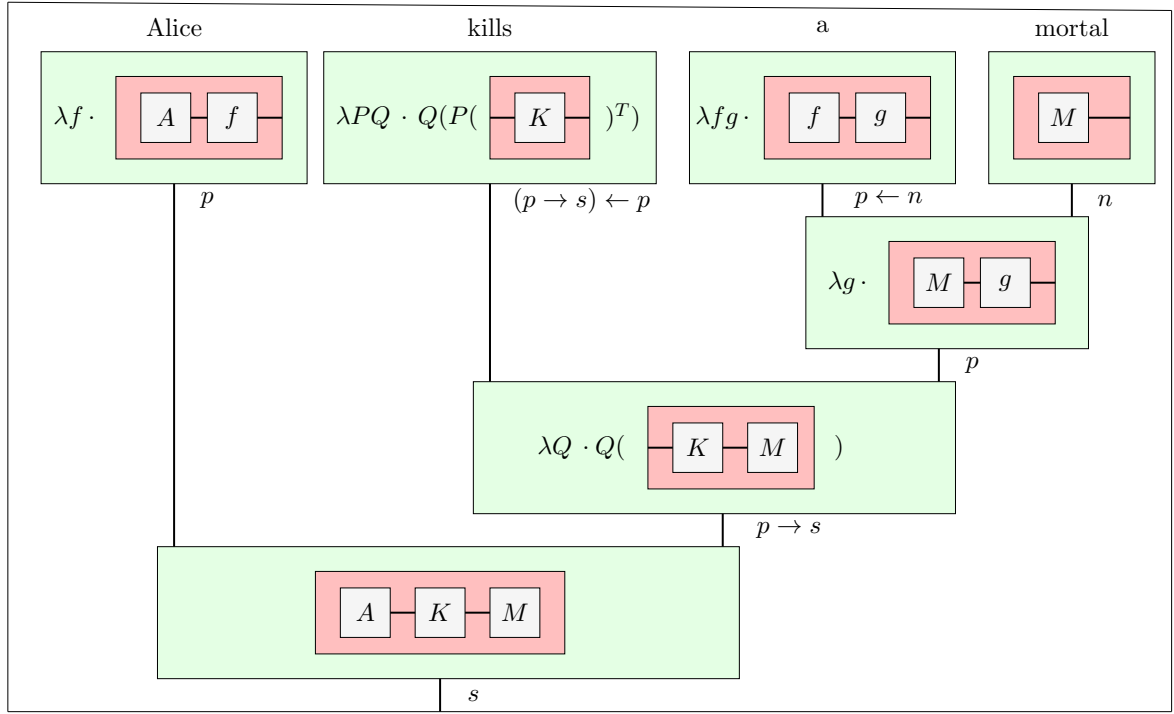

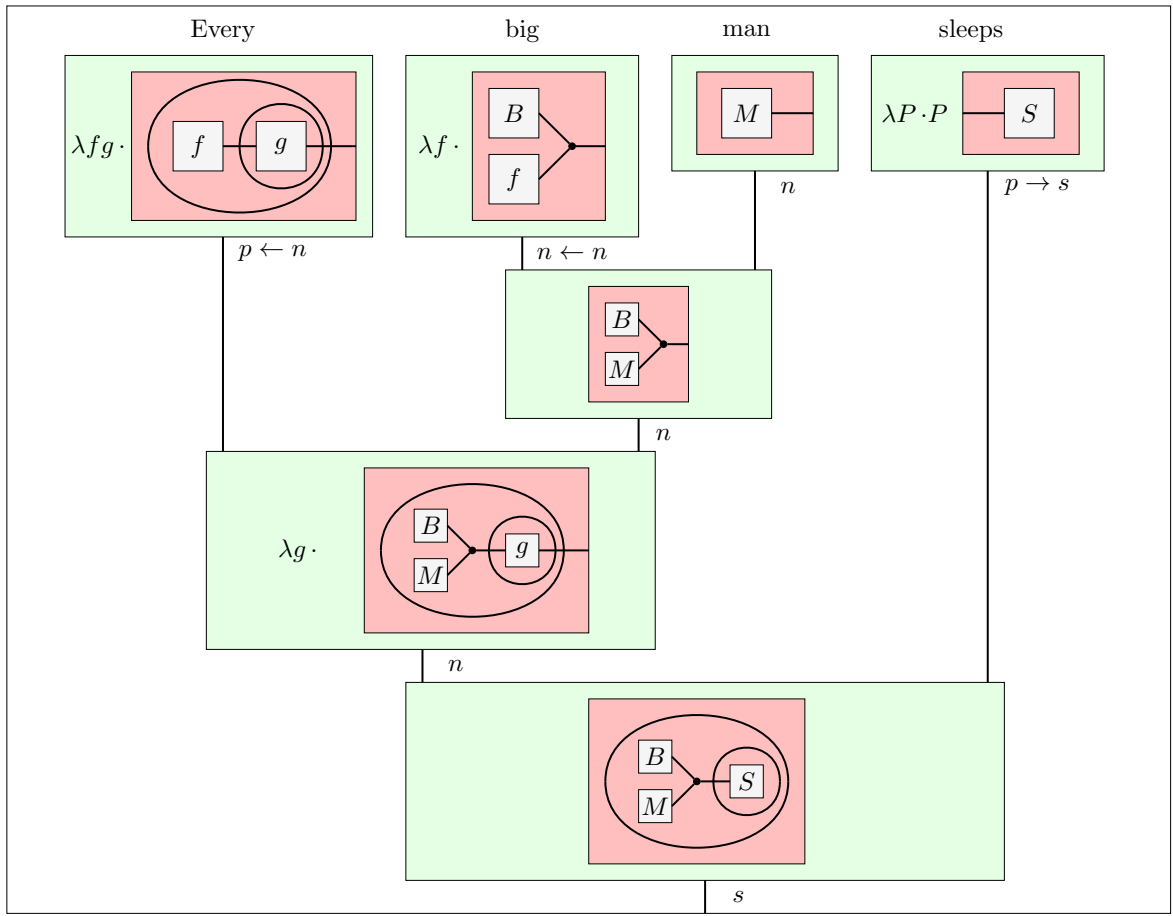
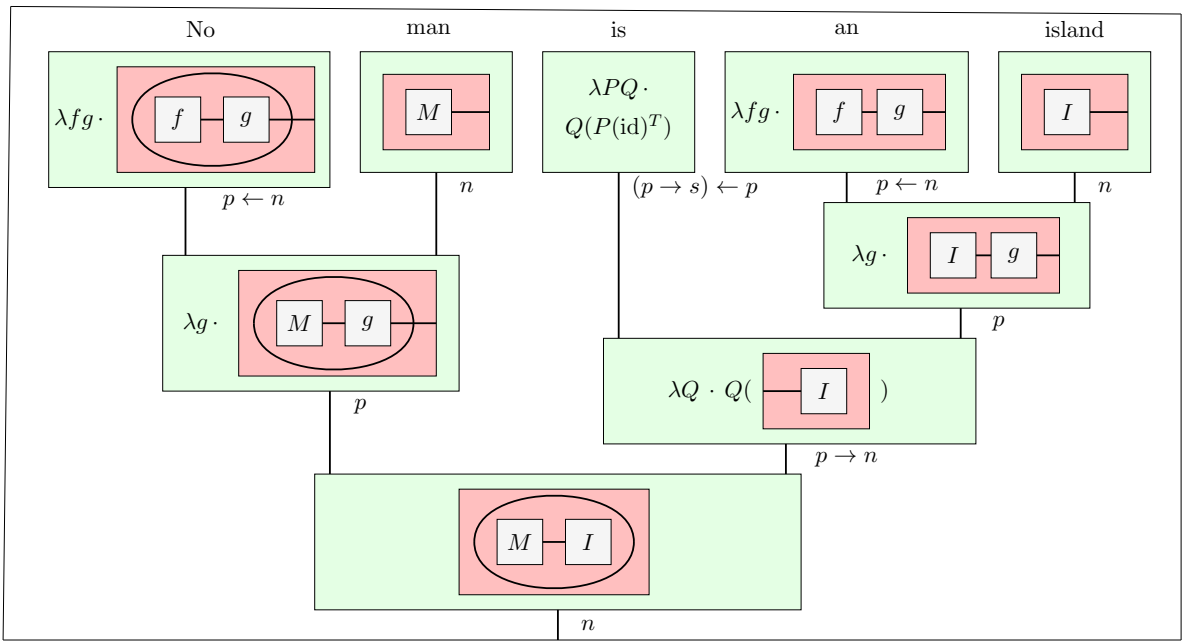

# Conclusion

Higher-order DisCoCat models are a generalisation of DisCoCat and Montague semantics where the meaning of words is given by lambda terms which output string diagrams. We have given a formal definition in terms of functors **G** → $\Lambda D$ for **G** a categorial grammar and $\Lambda D$ the cartesian closed category generated by the language of string diagrams. We defined a concrete instance of our model — *Peirce-Lambek-Montague semantics* — where the string diagrams are Peircean existential graphs. Using DisCoPy, [14] the Python library for applied category theory, the implementation of our formal definition is straightforward, see Appendix A. Giving experimental support for our higher-order DisCoCat models would be a natural next step. Another direction which has been developed recently [26] is to go from sentences to text, formalising the DisCoCirc framework [27] in terms of diagram-valued higher-order functions.

**Acknowledgements** The authors thank Sean Tull, Richie Yeung and Bob Coecke for their insightful support in improving this manuscript.

# References

[1] Stephen Clark, Bob Coecke, and Mehrnoosh Sadrzadeh. "A Compositional Distributional Model of Meaning". In: *Proceedings of the Second Symposium on Quantum Interaction (QI-2008)*. 2008, pp. 133–140.

[2] Stephen Clark, Bob Coecke, and Mehrnoosh Sadrzadeh. "Mathematical Foundations for a Compositional Distributional Model of Meaning". In: *A Festschrift for Jim Lambek*. Ed. by J. van Benthem and M. Moortgat. Vol. 36. Linguistic Analysis. 2010, pp. 345–384. arXiv: 1003.4394.

[3] Giovanni de Felice. "Categorical Tools for Natural Language Processing". PhD thesis. University of Oxford, Dec. 2022. arXiv: 2212.06636.

[4] Joachim Lambek. "The Mathematics of Sentence Structure". In: *The American Mathematical Monthly* 65.3 (Mar. 1958), pp. 154–170. DOI: 10.1080/00029890.1958.11989160.

[5] Bob Coecke, Edward Grefenstette, and Mehrnoosh Sadrzadeh. "Lambek vs. Lambek: Functorial Vector Space Semantics and String Diagrams for Lambek Calculus". In: *Ann. Pure Appl. Log.* 164.11 (2013), pp. 1079–1100. DOI: 10.1016/j.apal.2013.05.009. arXiv: 1302.0393.

[6] Richie Yeung and Dimitri Kartsaklis. "A CCG-Based Version of the DisCoCat Framework". In: *ArXiv e-prints* (May 2021). arXiv: 2105.07720.

[7] Giovanni de Felice, Konstantinos Meichanetzidis, and Alexis Toumi. "Functorial Question Answering". In: *Proceedings Applied Category Theory 2019, ACT 2019, University of Oxford, UK*. Vol. 323. EPTCS. 2019. DOI: 10.4204/EPTCS.323.6.

[8] Charles Santiago Sanders Peirce. "Prolegomena to an Apology of Pragmaticism". In: *The Monist* 16.4 (1906), pp. 492–546. JSTOR: 27899680.

[9] Geraldine Brady and Todd H. Trimble. "A String Diagram Calculus for Prediate Logic and C. S. Peirce's System Beta". 1998. URL: http://people.cs.uchicago.edu/~brady/beta98.ps.

[10] Nathan Haydon and Paweł Sobociński. "Compositional Diagrammatic First-Order Logic". In: *Diagrammatic Representation and Inference*. Ed. by Ahti-Veikko Pietarinen et al. Lecture Notes in Computer Science. Cham: Springer International Publishing, 2020, pp. 402–418. DOI: 10.1007/978-3-030-54249-8_32.

[11] Richard Montague. "Universal Grammar". In: *Theoria* 36.3 (1970), pp. 373–398. DOI: 10.1111/j.1755-2567.1970.tb00434.x.

[12] Bob Coecke, Giovanni de Felice, Konstantinos Meichanetzidis, and Alexis Toumi. "Foundations for Near-Term Quantum Natural Language Processing". In: *ArXiv e-prints* (2020). arXiv: 2012.03755.

[13] Alexis Toumi. "Category Theory for Quantum Natural Language Processing". PhD thesis. University of Oxford, 2022. arXiv: 2212.06615.

[14] Giovanni de Felice, Alexis Toumi, and Bob Coecke. "DisCoPy: Monoidal Categories in Python". In: *Proceedings of the 3rd Annual International Applied Category Theory Conference, ACT*. Vol. 333. EPTCS, 2020. DOI: 10.4204/EPTCS.333.13.

[15] Johan Bos et al. "Wide-Coverage Semantic Representations from a CCG Parser". In: *COLING 2004: Proceedings of the 20th International Conference on Computational Linguistics*. Geneva, Switzerland: COLING, Aug. 2004, pp. 1240–1246. URL: https://aclanthology.org/C04-1180 (visited on 11/29/2023).

[16] Samson Abramsky and Nikos Tzevelekos. "Introduction to Categories and Categorical Logic". In: *CoRR* abs/1102.1313 (2011). arXiv: 1102.1313.

[17] J. Lambek. "Categorial and Categorical Grammars". In: *Categorial Grammars and Natural Language Structures*. Ed. by Richard T. Oehrle, Emmon Bach, and Deirdre Wheeler. Studies in Linguistics and Philosophy. Dordrecht: Springer Netherlands, 1988, pp. 297–317. DOI: 10.1007/978-94-015-6878-4_11.

[18] Jan van Eijck and Christina Unger. *Computational Semantics with Functional Programming*. Cambridge: Cambridge University Press, 2010. DOI: 10.1017/CBO9780511778377.

[19] Peter Selinger. "A Survey of Graphical Languages for Monoidal Categories". In: *New Structures for Physics* (2010), pp. 289–355. DOI: 10.1007/978-3-642-12821-9_4.

[20] Giovanni de Felice, Elena Di Lavore, Mario Román, and Alexis Toumi. "Functorial Language Games for Question Answering". In: *Proceedings of the 3rd Annual International Applied Category Theory Conference 2020, ACT 2020, Cambridge, USA, 6-10th July 2020*. Ed. by David I. Spivak and Jamie Vicary. Vol. 333. EPTCS. 2020, pp. 311–321. DOI: 10.4204/EPTCS.333.21.

[21] Roger Penrose and Wolfgang Rindler. *Spinors and Space-Time: Volume 1: Two-Spinor Calculus and Relativistic Fields*. Vol. 1. Cambridge Monographs on Mathematical Physics. Cambridge: Cambridge University Press, 1984. DOI: 10.1017/CBO9780511564048.

[22] G. Chiribella, G. M. D'Ariano, and P. Perinotti. "Quantum Circuit Architecture". In: *Physical Review Letters* 101.6 (Aug. 2008), p. 060401. DOI: 10.1103/PhysRevLett.101.060401.

[23] Mario Román. "Coend Calculus and Open Diagrams". In: *ArXiv e-prints* (Apr. 2020). arXiv: 2004.04526.

[24] James Hefford and Cole Comfort. "Coend Optics for Quantum Combs". In: *Electronic Proceedings in Theoretical Computer Science* 380 (Aug. 2023), pp. 63–76. DOI: 10.4204/EPTCS.380.4. arXiv: 2205.09027 [quant-ph].

[25] Donald Yau. *Operads of Wiring Diagrams*. Vol. 2192. Lecture Notes in Mathematics. Cham: Springer International Publishing, 2018. DOI: 10.1007/978-3-319-95001-3.

[26] Jonathon Liu et al. "Discourse Text Circuits From CCG: A Pipeline". In: *to appear* (2023).

[27] Bob Coecke. "The Mathematics of Text Structure". In: *Joachim Lambek: The Interplay of Mathematics, Logic, and Linguistics*. Ed. by Claudia Casadio and Philip J. Scott. Cham: Springer International Publishing, 2021, pp. 181–217. DOI: 10.1007/978-3-030-66545-6_6.

# A   Implementation

```
# Step 1: Define Formula as a subclass of frobenius.Diagram

from discopy import frobenius
from discopy.tensor import Dim, Tensor
from discopy.cat import Category, factory

@factory
class Formula(frobenius.Diagram):
    ty_factory = frobenius.PRO  # i.e. natural numbers as objects

    def eval(self, size):
        return frobenius.Functor(
            ob=lambda _: Dim(size),
            ar=lambda box: box.data,
            cod=Category(Dim, Tensor[bool]))(self)

class Cut(frobenius.Bubble, Formula): pass
class Ligature(frobenius.Spider, Formula): pass
class Predicate(frobenius.Box, Formula): pass

Id, Formula.bubble_factory = Formula.id, Cut
Tensor[bool].bubble = lambda self, **_: self.map(lambda x: not x)
```

```
# Step 2: Parse natural language sentences using a categorial grammar

from discopy.grammar.categorial import Ty, Word, Eval

n, p, s = Ty('n'), Ty('p'), Ty('s')  # noun, phrase and sentence

Alice = Word("Alice", p)
big, sleeps = Word("big", n << n), Word("sleeps", p >> s)
man, island = (Word(noun, n) for noun in ("man", "island"))
kills, _is = (Word(verb, (p >> s) << p) for verb in ("kills", "is"))
no, every, some = (Word(det, p << n) for det in ("no", "every", "some"))

Alice_kills_a_mortal = (Alice @ kills @ some @ man
    >> p @ ((p >> s) << p) @ Eval(p << n)
    >> p @ Eval((p >> s) << p) >> Eval(p >> s))
every_big_man_sleeps = (every @ big @ man @ sleeps
    >> ((p << n) @ Eval(n << n) >> Eval(p << n))
    @ (p >> s) >> Eval(p >> s))
no_man_is_an_island = (no @ man @ _is @ some @ island
    >> Eval(p << n) @ ((p >> s) << p) @ Eval(p << n)
    >> p @ Eval((p >> s) << p) >> Eval(p >> s))
```

```python
from random import choice

size = 42  # Generating a random interpretation to test our model
random_bits = lambda n=size: [choice([True, False]) for _ in range(n)]

is_killed_by = [random_bits() for _ in range(size)]
unary_predicates = is_Alice, is_man, is_island, is_big, is_sleeping = [
    random_bits() for _ in range(5)]

K = Predicate("K", 1, 1, data=is_killed_by)
A, M, I, B, S = (Predicate(P, 0, 1, data)
                 for P, data in zip("AMIBS", unary_predicates))
```

```python
# Step 3: Higher-order DisCoCat as a closed functor into Python functions

from typing import Callable

from discopy import closed
from discopy.python import Function

F = closed.Functor(
    cod=Category(tuple[type, ...], Function),
    ob={s: Formula, n: Formula, p: Callable[[Formula], Formula]},
    ar={Alice: lambda: lambda f: A >> f,
        sleeps: lambda: lambda P: P(S.dagger()),
        man: lambda: M, island: lambda: I,
        big: lambda: lambda f: f @ B >> Ligature(2, 1, frobenius.PRO(1)),
        _is: lambda: lambda P: lambda Q: Q(P(Id(1)).dagger()),
        kills: lambda: lambda P: lambda Q: Q(P(K).dagger()),
        no: lambda: lambda f: lambda g: (f >> g).bubble(),
        some: lambda: lambda f: lambda g: f >> g,
        every: lambda: lambda f: lambda g: (f >> g.bubble()).bubble()})

evaluate = lambda sentence: bool(F(sentence)().eval(size))

assert evaluate(Alice_kills_a_mortal) == any(
    is_man[y] and is_killed_by[y][x] and is_Alice[x]
    for x in range(size) for y in range(size))
assert evaluate(every_big_man_sleeps) == all(
    not (is_big[x] and is_man[x]) or is_sleeping[x] for x in range(size))
assert evaluate(no_man_is_an_island) == all(
    not is_man[x] or not is_island[x] for x in range(size))
```

This implementation is also available as a notebook:

https://docs.discopy.org/en/main/notebooks/higher-order-discocat.html